# 基于全局融合的多核概念分解算法

李 飞[1,2],杜 亮[1,2,3*],任超宏[1,2]

(1.山西大学 计算机与信息技术学院,太原 030006; 2.山西大学 大数据科学与产业研究院,太原 030006;
3.计算智能与中文信息处理教育部重点实验室(山西大学),太原 030006)
(*通信作者电子邮箱 im.duliang@qq.com)

**摘 要:**非负矩阵分解(NMF)算法仅能用于对原始非负数据寻找低秩近似,而概念分解(CF)算法将矩阵分解模型扩展到单个非线性核空间,提升了矩阵分解算法的学习能力和普适性。针对无监督环境下概念分解面临的如何设计或选择合适核函数这一问题,提出基于全局融合的多核概念分解(GMKCF)算法。同时输入多种候选核函数,在概念分解框架下基于全局线性权重融合对它们进行学习,以得出质量高稳定性好的聚类结果,并解决概念分解模型面临核函数选择的问题。采用交替迭代的方法对新模型进行求解,证明了算法的收敛性。将该算法与基于核的*K*-均值(KKM)、谱聚类(SC)、KCF(Kernel Concept Factorization)、Coreg(Co-regularized multi-view spectral clustering)、RMKKM(Robust Multiple KKM)在多个真实数据库上的实验结果表明,该算法在数据聚类方面优于对比算法。

# Multiple kernel concept factorization algorithm based on global fusion

LI Fei[1,2], DU Liang[1,2,3*], REN Chaohong[1,2]

(1. *School of Computer and Information Technology*, *Shanxi University*, *Taiyuan Shanxi* 030006, *China*;
2. *Institute of Big Data Science and Industry*, *Shanxi University*, *Taiyuan Shanxi* 030006, *China*;
3. *Key Laboratory of Computational Intelligence and Chinese Information Processing*, *Ministry of Education*
(*Shanxi University*), *Taiyuan Shanxi* 030006, *China*)

**Abstract:** Non-negative Matrix Factorization (NMF) algorithm can only be used to find low rank approximation of original non-negative data while Concept Factorization (CF) algorithm extends matrix factorization to single non-linear kernel space, improving learning ability and adaptability of matrix factorization. In unsupervised environment, to design or select proper kernel function for specific dataset, a new algorithm called Globalized Multiple Kernel CF (GMKCF) was proposed. Multiple candidate kernel functions were input in the same time and learned in the CF framework based on global linear fusion, obtaining a clustering result with high quality and stability and solving the problem of kernel function selection that the CF faced. The convergence of the proposed algorithm was verified by solving the model with alternate iteration. The experimental results on several real databases show that the proposed algorithm outperforms comparison algorithms in data clustering, such as Kernel *K*-Means (KKM), Spectral Clustering (SC), Kernel CF (KCF), Co-regularized multi-view spectral clustering (Coreg), and Robust Multiple KKM (RMKKM).

## 0 引言

数据挖掘从看似无序的数据中寻找有序、有价值的信息。聚类分析是数据挖掘、机器学习中的一项重要技术,也是国内外学者研究的一个重点领域。聚类技术可用来探索数据的内部结构,并就其某种相关关系进行挖掘,因而在很多领域中得到广泛应用,例如:在电子商务中,应用聚类算法可以发现不同客户群体,有利于寻找潜在市场;在生物学领域,可以对基因、蛋白质等进行聚类研究,从而获取对其结构的深入认识;在互联网上,可以对微博、新闻中的文档进行聚类研究,从而进行热点事件发现等。

根据聚类算法的输入数据类型分类,聚类算法可以分为数值型算法(如*K*-means[1]、非负矩阵分解(Non-negative Matrix Factorization,NMF)[2]等)、离散型(如AT-DC[3])、关系型(如NCut[4]、仿射传播AP[5])和混合型(如图正则的NMF(Graph regularized NMF,GNMF)[6])算法等。根据输出结果,聚类算法分为层次聚类和划分式聚类[1]。根据簇的描述形式,聚类算法可分为基于原型的方法(也叫簇代表元,代表性算法有*K*-means,*K*-medoids等)和基于模型的方法(代表性算法有高斯混合模型(Gaussian Mixture Model,GMM)[7])。

近年来研究人员提出许多方法进一步提高非负矩阵分解算法的效果。文献[6]提出利用数据流形结构提升聚类结

**收稿日期:**2018-09-13;**修回日期:**2018-11-21;**录用日期:**2018-11-23。 **基金项目:**国家自然科学基金资助项目(61502289)。

**作者简介:**李飞(1993—),男,湖北黄冈人,硕士研究生,CCF会员,主要研究方向:数据挖掘、机器学习; 杜亮(1985—),男,山西晋中人,讲师,博士,CCF会员,主要研究方向:数据挖掘、机器学习; 任超宏(1994—),男,山西朔州人,硕士研究生,主要研究方向:数据挖掘、机器学习。

果; 文献[8]研究矩阵分解的稀疏性提高结果的可解释性; 文献[9]研究噪声数据上的矩阵分解提高分解结果的鲁棒性。文献[10]提出概念分解(Concept Factorization, CF)方法将矩阵分解从线性原始空间扩展到非线性核空间并用于文本聚类,文献[11]提出基于图正则的概念分解算法,文献[12]进一步提出自适应邻居正则化的概念分解算法,文献[13-15]提出基于多图/多层/多视图的正则化的概念分解算法,文献[16-18]提出新型单视图数据的正则化概念分解算法。值得指出的是这类正则化方法通常需要引入额外的参数用于平衡概念分解目标函数和正则目标,但实际应用中如何设置较为准确的参数是比较困难的。

文献[10]提出的核概念分解方法在实际应用中面临的核心问题之一是针对特定任务和数据集如何设计和选择合适的核函数。需要进一步指出的是,由于缺乏数据标签等监督信息,核函数选择在无监督学习任务中变得更加困难。

为了减轻核函数选择带来的困难,本文提出在无监督多核学习框架中通过全局线性加权方法从一系列初始给定的核矩阵中学习聚类质量更高、稳定性更好的核函数。针对本文提出的多核概念分解模型,推导和设计了对应的迭代式优化求解算法——基于全局融合的多核概念分解(Globalized Multiple Kernel CF, GMKCF)算法,并证明该算法的收敛性以及算法的时间和空间复杂度。本文提出的多核概念分解模型没有引入额外的超参数,降低了算法在实际应用中实施部署的难度。

多个基准数据集上的聚类实验结果表明多核聚类方法明显优于单核平均结果,验证了多核学习可以提升聚类算法性能。本文提出的多核概念分解在聚类准确性、归一化互信息和聚类纯度上的性能优于对比多核聚类方法。

## 1 相关工作

### 1.1 非负矩阵分解

针对非负值矩阵数据 $X \in \mathbf{R}^{d\times n}$, Lee 等[2]于 1999 年在《Nature》上正式提出了非负矩阵分解的基本概念。非负矩阵分解的认知基础是: 对整体的感知由基于对组成整体的部分(局部)。NMF 通过非负约束纯加性的感知过程刻画出数据的组成部分和数据如何由局部感知构成的本质。该方法用两个低秩非负矩的乘积阵 $UV^T$ 近似原始非负数据,其中 $U \in \mathbf{R}^{d\times k}$, $V \in \mathbf{R}^{n\times k}$。非负矩阵分解方法对应的最优结果可以通过求解以下优化问题[19]获得:

$$\min_{U,V} \| X - UV^{\mathrm{T}} \|^2;\ U \geqslant 0, V \geqslant 0 \tag{1}$$

从式(1)可看出: 每一个样本 $x_i$ 可以通过 $U,V$ 的线性合并得到,即 $x_i = \sum_k u_k v_{ik}$。因此,矩阵 $U$ 可以看作是一组非负基向量,而矩阵 $V$ 可以看作数据在基矩阵 $U$ 下新的表示。

上述优化问题是关于联合$(U,V)$的非凸优化问题,因此很难用非线性优化方法得到全局最优解。然而对于仅关于 $U$ 或者仅关于 $V$ 的子问题,仍然是一个凸优化问题。其局部最优解可以通过分块坐标轮换法分别求解。通用的非负矩阵分解求解算法通过以下乘法更新公式获得:

$$\begin{cases} U_{ij} = U_{ij}\dfrac{(XV)_{ij}}{(UV^{\mathrm{T}}V)_{ij}} \\ V_{ij} = V_{ij}\dfrac{(X^{\mathrm{T}}U)_{ij}}{(VU^{\mathrm{T}}U)_{ij}} \end{cases} \tag{2}$$

### 1.2 概念分解

Xu 等在文献[10]中提出概念分解算法。在概念分解模型中,分解后的基向量 $u$ 要求通过对原始空间样本的非负线性组合得到,其对应的优化问题可以写成:

$$\min_{U,V} \| X - XUV^{\mathrm{T}} \|^2;\ U \geqslant 0, V \geqslant 0$$

文献[2]中提出的非负矩阵分解方法仅适用于原始特征。对于高度非线性分布的数据集,可以利用核方法来提降矩阵分解结果。核函数是从低维空间到高维空间的一种映射函数对于输入空间 $x \in \mathbf{R}^{d\times 1}$,函数 $\varphi(x)$ 将输入空间映射到希尔伯特特征空间 $H$。对于 $x \in \mathbf{R}^{d\times 1}$, $y \in \mathbf{R}^{d\times 1}$,函数 $k(x, y)$ 称之为核函数的条件是满足以下条件:

$$k(x, y) = \varphi(x)^{\mathrm{T}}\varphi(y)$$

映射函数 $\varphi(x)$ 将输入投影到高维空间后引发难以计算的问题,核方法的关键之一是核技巧(kernel trick)。核技巧将高维空间的计算问题转化为低维度核函数计算问题。通常满足 Mercer 定理[7],即定义任何半正定的函数都可以作为核函数。常用的核函数有: 线性核、多项式核、高斯核等,分别定义如下:

$$k(x, y) = x^{\mathrm{T}}y$$

$$k(x, y) = (1 + x^{\mathrm{T}}y)^d$$

$$k(x, y) = \exp\left(-\frac{\| x - y \|^2}{2\delta^2}\right)^d$$

为此,文献[10]中进一步将概念分解模型扩展到核空间,核概念分解对应的优化问题变为:

$$\min_{U,V} \mathrm{tr}(K) - 2\mathrm{tr}(V^{\mathrm{T}}KU) + \mathrm{tr}(U^{\mathrm{T}}KUV^{\mathrm{T}}V) \tag{3}$$

s. t. $U \geqslant 0, V \geqslant 0$

对于非负核矩阵,上述优化问题的最优解可以通过下面的乘法更新公式获得:

$$\begin{cases} U_{ij} = U_{ij}\dfrac{(KV)_{ij}}{(KUV^{\mathrm{T}}V)_{ij}} \\ V_{ij} = V_{ij}\dfrac{(KU)_{ij}}{(VU^{\mathrm{T}}KU)_{ij}} \end{cases} \tag{4}$$

## 2 本文算法

上述提到的核概念分解算法仅适用于单核数据聚类问题; 然而,核方法面临的核心问题之一是针对特定任务和数据集如何设计和选择合适的核函数。需要进一步指出的是,由于缺乏数据标签等监督信息,核函数选择在无监督学习任务中变得更加困难。

为了减轻核函数选择带来的困难,本文提出在无监督多核学习框架中通过全局线性加权方法从一系列初始给定的核矩阵中学习聚类质量更高、稳定性更好的核函数。

### 2.1 多核概念分解

假设一共给定 $m$ 个不同的核关系数据用于聚类过程 $\{K^i\}_{i=1}^m$,与此对应的是 $m$ 个不同的特征空间$\{H^i\}_{i=1}^m$。为了合并这些核空间并且使得合并后的核矩阵仍然满足 Mercer 条件,可以采用基于非负全局权重线性加权的方式,即合并后的特征空间可以表示为:

$$\varphi'(x) = \sum_{i=1}^{m} w_i\varphi_i(x);\ w_i \geqslant 0 \tag{5}$$

然而需要指出的是上述方法并不可行,原因在于不同的

特征空间 $\varphi_i$ 对应的维度并不相同。因此,本文通过将不同的核函数对应的特征空间加权拼接起来得到一个扩张的高维映射 $\varphi_w(x) = [w_1\varphi_1(x); w_2\varphi_2(x); \cdots; w_m\varphi_m(x)]$,进而构造一个增广希尔伯特空间 $H_w$。

现有研究工作表明半正定核矩阵 $\{K^i\}_{i=1}^m$ 的非负凸组合

$$K_w = \sum_{i=1}^m K^i;\ w_i \geqslant 0 \tag{6}$$

仍然是一个半正定核矩阵。通过替换核概念分解中的核矩阵为新的多核矩阵,就可以得到 GMKCF 算法,对应的优化问题可以表示为:

$$\min \operatorname{tr}(K_w) - 2\operatorname{tr}(V^T K_w U) + \operatorname{tr}(U^T K_w U V^T V) \tag{7}$$

$$\text{s.t.}\quad U \geqslant 0, V \geqslant 0,\ \sum_{i=1}^m w_i = 1;\ w_i \geqslant 0$$

2.2　多核概念分解模型求解算法

首先需要指出的是,上述多核概念分解模型整体对于所有待求变量仍然是一个非凸优化问题,但是对于单个变量的各子优化问题都是凸优化问题。为此,本文提出迭代式求解算法对整体问题进行求解,并采用分块坐标轮换法分别对每个变量对应的子优化问题进行单独求解。最终通过求解一系列子优化问题,可以获得对应的局部最优解。

2.2.1　固定 $V$、$w$,求解 $U$

多核概念分解关于 $U$ 的子优化问题变成:

$$\min_U \operatorname{tr}(U^T K_w U V^T V) - 2\operatorname{tr}(V^T K_w U) \tag{8}$$

$$\text{s.t.}\quad U \geqslant 0$$

容易看出,上述优化问题等价于采用 $K_w$ 的单核概念分解模型对应的子优化问题。

因此,上述问题的最优解同样可以通过下述乘法更新公式获得:

$$U_{ij} = U_{ij}\frac{(K_w V)_{ij}}{(K_w U V^T V)_{ij}} \tag{9}$$

对于 $K_w$ 中有负数的情况,对应的更新公式变为:

$$U_{ij} = U_{ij}\frac{(K_w V)_{ij}^2 + \sqrt{(K_w V)_{ij} + 4(P_w^+)_{ij}(P_w^-)_{ij}}}{2(P_w^+)_{ij}} \tag{10}$$

$P_w^+$、$P_w^-$ 的计算公式为:

$$\begin{cases} K_w^+ = \dfrac{|K_w| + K_w}{2} \\ K_w^- = \dfrac{|K_w| - K_w}{2} \\ P_w^+ = K_w^+ U V^T V \\ P_w^- = K_w^+ U V^T V \end{cases} \tag{11}$$

2.2.2　固定 $U$、$w$,求解 $V$

多核概念分解关于 $V$ 的子优化问题变成:

$$\min_V \operatorname{tr}(U^T K_w U V^T V) - 2\operatorname{tr}(V^T K_w U) \tag{12}$$

$$\text{s.t.}\quad V \geqslant 0$$

容易看出,上述优化问题等价于采用 $K_w$ 的单核概念分解模型对应的子优化问题。

因此,上述问题的最优解同样可以通过下述乘法更新公式获得:

$$V_{ij} = V_{ij}\frac{(K_w U)_{ij}}{(V U^T K_w U)_{ij}} \tag{13}$$

对于 $K_w$ 中有负数的情况,对应的更新公式变为:

$$V_{ij} = V_{ij}\frac{(K_w U)_{ij}^2 + \sqrt{(K_w U)_{ij} + 4(Q_w^+)_{ij}(Q_w^-)_{ij}}}{2(Q_w^+)_{ij}} \tag{14}$$

$Q_w^+$、$Q_w^-$ 的计算公式为:

$$\begin{cases} Q_w^+ = V U^T K_w^+ V \\ Q_w^- = V U^T K_w^- V \end{cases} \tag{15}$$

2.2.3　固定 $U$、$V$,求解 $w$

首先定义 $e \in \mathbf{R}^{m\times 1}$,且令

$$e_i = \operatorname{tr}(K^i) - 2\operatorname{tr}(V^T K^i U) + \operatorname{tr}(U^T K^i U V^T V)$$

多核概念分解关于 $w$ 的子优化问题变成:

$$\min_w \sum_{i=1}^m w_i^2 e_i \tag{16}$$

$$\text{s.t.}\quad \sum_{i=1}^m w_i = 1;\ w_i \geqslant 0$$

上述问题关于变量 $w$ 的拉格朗日扩展可以写为:

$$J(w) = \sum_{i=1}^m w_i^2 e_i + \lambda\left(1 - \sum_{i=1}^m w_i\right)$$

该目标最优解对应的 KKT 条件 $\frac{\partial J(w)}{\partial w} = 0$,且需要满足等式约束 $\sum_{i=1}^m w_i = 1$。通过进一步求解,可以获得关于变量 $w$ 的闭式解:

$$w_i = \frac{\dfrac{1}{e_i}}{\sum_{j=1}^m \dfrac{1}{e_j}} \tag{17}$$

2.2.4　多核概念算法

算法 1　全局多核概念分解算法。

输入　样本-特征关系矩阵 $X \in \mathbf{R}^{d\times n}$,概念因子个数 $k$。

输出　多核全局意义下样本的低秩表示 $V \in \mathbf{R}^{n\times k}$。

1) 采用多个核函数($m$)分别得到多个对应的核矩阵 $\{K^i\}_{i=1}^m$;
2) 初始化非负因子 $U$ 和 $V$;
3) 初始化核函数权重因子 $w_i = \frac{1}{m}$;
4) 计算初始目标函数 $obj_{\text{old}}$;
5) While not converge do
6) 　根据式(6)计算 $K_w$;
7) 　根据式(9)或式(10)计算 $U$;
8) 　根据式(13)或式(14)计算 $V$;
9) 　根据式(17)计算 $w$;
10) 　计算当前目标函数 $obj_{\text{new}}$;
11) 　如果 $\frac{obj_{\text{old}} - obj_{\text{new}}}{obj_{\text{new}}} \leqslant 10^{-5}$,判定收敛,终止迭代;
12) End While

后处理: 利用 $K$-means 算法对多核低秩表示 $V$ 进行二次聚类获得高质量的离散化聚类结果。

2.2.5　算法收敛性证明

式(7)中的全局多核概念分解算法是一个关于联合 $(\{U^i\}_{i=1}^m, V, w)$ 的非凸优化问题,因此很难用非线性优化方法得到全局最优解。然而对于仅关于 $\{U^i\}_{i=1}^m$ 或者仅关于 $V$ 或者仅关于 $w$ 的子问题,仍然是一个凸优化问题。通过分块坐标轮换法的迭代式求解可以使整体目标函数单调下降。并且可以很明显看出式(7)的目标函数是有下界的。因此,整体求

解算法的收敛性可以得到保障。

具体来讲，容易看出式(7)的目标函数是有下界的(下界为0)，并且式(7)的函数值随着算法迭代每一步都是非增的(Non-increasing)。本文引入非负矩阵分解(NMF)和概念分解(CF)模型乘法更新过程(Multiplicative update rule)中常见的辅助函数(Auxiliary function)定义[10]。因为非负因子 $\boldsymbol{U}$ 的更新过程和非负因子 $\boldsymbol{V}$ 更新类似，本文仅给出求解非负因子 $\boldsymbol{V}$ 时的辅助函数证明。

**定义1** 满足以下条件的函数 $G(v,v')$ 是函数 $F(v)$ 的辅助函数：$G(v,v') \geqslant F(v)$，$G(v,v) = F(v)$。

**引理1** 如果函数 $G$ 是函数 $F$ 的辅助函数，用 $v^{t+1} = \arg\min_v G(v,v^t)$ 更新公式进行更新时函数 $F$ 是非增的。

**证明** $F(v^{t+1}) \leqslant G(v^{t+1},v^t) \leqslant G(v^t,v^t) = F(v^t)$。

式(12)是关于整个非负因子 $\boldsymbol{V}$ 的优化问题。简单起见，本文对元素 $v_{ab}$ 对应的优化问题进行分析。令与元素 $v_{ab}$ 相关的目标函数为 $F_{ab}$，则其导数和二阶导数如下：

$$F'_{ab} = (-2\boldsymbol{K}_w\boldsymbol{U} + 2\boldsymbol{V}\boldsymbol{U}^{\mathrm{T}}\boldsymbol{K}_w\boldsymbol{U})_{ab}$$

$$F''_{ab} = 2(\boldsymbol{U}^{\mathrm{T}}\boldsymbol{K}_w\boldsymbol{U})_{ab}$$

**引理2** 函数 $F_{ab}(v)$ 的辅助如下：

$$G(v,v^t_{ab}) = F_{ab}(v^t_{ab}) + F'_{ab}(v^t_{ab})(v - v^t_{ab}) + \frac{(\boldsymbol{VUKwU})_{ab}}{v^t_{ab}}(v - v^t_{ab})^2$$

**证明** 显然 $G(v,v) = F_{ab}(v)$。$F_{ab}(v)$ 的二阶泰勒展开如下：

$$F_{ab}(v) = F_{ab}(v^t_{ab}) + F'_{ab}(v^t_{ab})(v - v^t_{ab}) + (\boldsymbol{VUKwU})_{ab}(v - v^t_{ab})^2$$

此外，$(\boldsymbol{VUKwU})_{ab} = \sum_{l=1}^{k} v^t_{al}(\boldsymbol{UKwU})_{lb} \geqslant v^t_{ab}(\boldsymbol{UKwU})_{ab}$。因此 $G(v,v^t_{ab}) \geqslant F_{ab}(v)$。

辅助函数 $G(v,v^t_{ab})$ 是简单的一元二次函数，将其引入定义1并求解 $\min_v G(v,v^t_{ab})$ 可以得到：

$$v^{t+1}_{ab} = v^t_{ab} - v^t_{ab}\frac{F'_{ab}(v^t_{ab})}{2(\boldsymbol{VUK}_w\boldsymbol{U})_{ab}} = v^t_{ab}\frac{(\boldsymbol{K}_w\boldsymbol{U})_{ab}}{(\boldsymbol{VUK}_w\boldsymbol{U})_{ab}}$$

更新式(13)得证。

此外，式(16)中关于 $\boldsymbol{w}$ 的问题是凸优化问题，式(17)可以获得最优解。

### 2.2.6 算法复杂性说明

初始阶段，本文算法需要计算 $m$ 个核矩阵，对应的计算复杂性是 $O(mn^2d)$，其中 $n$ 是样本个数，$d$ 是特征个数。每次迭代过程中的计算量分别为：

1) 更新变量 $\boldsymbol{U}$，其中需要计算 $P^+$ 和 $P^-$，对应的计算复杂性为 $O(n^2k + k^2n)$，更新 $\boldsymbol{U}$ 的复杂性是 $O(n^2k)$。

2) 更新变量 $\boldsymbol{V}$，其中需要计算 $Q^+$ 和 $Q^-$，对应的计算复杂性为 $O(n^2k + k^2n)$，更新 $\boldsymbol{V}$ 的复杂性是 $O(n^2k)$。

3) 更新变量 $\boldsymbol{w}$，对应的计算复杂性为 $O(m(n^2k + k^2n))$。

4) 更新变量 $\boldsymbol{K}_w$，对应的计算复杂性为 $O(mn^2)$。

假设迭代算法在迭代 $t$ 次后收敛，多核概念分解的整体复杂度表示为 $O(mn^2d + n^2t(k+m))$。可以看出，多核概念分解整体算法复杂性和单核概念分解在同一量级。

## 3 实验与结果分析

本文实验通过基准测试数据集上的聚类结果对比来验证本文提出的多核方法在聚类问题上的有效性。

实验平台的配置：PC 为 Intel Core i5 处理器，8 GB 内存，120 GB 硬盘；操作系统为 Windows 10；编程环境为 Matlab 2015a。

### 3.1 数据集的选择

本文分别选择了 BBC、TR31、K1B、WebKB 四个数据集作为测试基准数据集。这些数据集经常被用于评估聚类算法的性能，数据集的统计信息如表1所示。

BBC 数据集包含了来自 BBC 新闻网站提供的 2 225 份文件，对应于 2004—2005 年 5 个主题领域的故事，共有 5 类标签：商业、娱乐、政治、体育、科技。

TR31 数据集来自 TREC 收集的文本数据集，包含 927 个文本，分为 7 个类别。

K1B 数据集来自 WebACE 项目，包括 2 340 篇文章，这些文章来自于路透新闻的 20 个类别中，其中每个文档对应于 Yahoo! 的主题层次结构中列出的网页。

WebKB 数据集包含了约 6 000 个从 4 所高校(康奈尔大学、德克萨斯大学、华盛顿大学、威斯康星大学)的计算机科学部门收集的网页。每个网页都标有一个标签：学生、教授、课程、项目、人员、部门，以及其他。

表1 实验中使用的数据集

Tab. 1 Datasets used in the experiment

| 数据集名 | 类数 | 实例数 | 维数 |
|---|---|---|---|
| BBC | 5 | 737 | 1 000 |
| TR31 | 7 | 927 | 10 128 |
| K1B | 6 | 2 340 | 21 839 |
| WebKB | 4 | 4 199 | 1 000 |

和其他多核学习方法中的策略类似，本文使用了 12 种不同的核函数作为多核学习的输入。这些核函数包括 7 个不同带宽的径向基函数(Radial Basis Function，RBF)核函数 $k(\boldsymbol{x},\boldsymbol{y}) = \exp\left(-\frac{\|\boldsymbol{x}-\boldsymbol{y}\|^2}{2\delta^2}\right)^d$，其中令 $\delta = tD_0$，且 $D_0$ 是样本两两之间距离的平均值，而 $t$ 的变化范围包括{0.01，0.05，0.1，1，10，50，100}；4 个多项式核函数 $k(\boldsymbol{x},\boldsymbol{y}) = (a + \boldsymbol{x}_i^{\mathrm{T}}\boldsymbol{y})^b$，其中 $a$ 的取值范围包括{0，1}，$b$ 的取值范围包括{2，4}；1 个余弦核函数 $k(\boldsymbol{x},\boldsymbol{y}) = \frac{\boldsymbol{x}^{\mathrm{T}}\boldsymbol{y}}{\|\boldsymbol{x}\| \cdot \|\boldsymbol{y}\|}$。最后，所有的核函数都又经过了标准化 $k(\boldsymbol{x},\boldsymbol{y}) = \frac{k(\boldsymbol{x},\boldsymbol{y})}{\sqrt{k(\boldsymbol{x},\boldsymbol{x})k(\boldsymbol{y},\boldsymbol{y})}}$，并且被进一步缩放到区间[0,1]内。

### 3.2 对比方法

本文实验是多核数据聚类实验，实验中对比了单核方法和多核方法。采用的单核方法包括：基于核的 $K$-均值(Kernel $K$-Means，KKM)、谱聚类(Spectual Clustering，SC)、KCF(Kernel CF)。采用的多核方法包括：Coreg(Co-regularized multi-view spectral clustering)[20]、RMKKM (Robust Multiple KKM)[21],以及本文 GMKCF 算法。

针对多核实验数据,单核方法可以获得多组实验结果,为了准确刻画单核方法在不同核函数上的性能,本文实验采用单核方法在多个核函数上聚类结果的平均值。

根据文献[20-21]中的实验结果,Coreg 在本文实验中的参数设置为 0.1,RMKKM 的实验参数设置为 0.3。概念因子的个数设置为数据集中类的个数。

聚类中簇的个数设置为数据集中真实类别的个数。SC 和 Coreg 获得样本低维表示后都采用 $K$-means 算法最终得到离散化的聚类结果。针对聚类算法需要初始化的问题,本文实验采用随机值对算法进行初始化,重复实验 20 次并报告对应的平均值。

### 3.3 评价指标

因本文实验所采用的数据集类别标签已知,本文选择三个外部评价指标来评估算法在聚类问题上的性能,各评价指标介绍如下:

1) 聚类准确性(Clustering Accuracy, ACC)。聚类准确性是基于类和簇的一一对应关系来评价聚类性能,对于样本 $x_i$,$p_i$ 和 $q_i$ 分别为聚类结果和真实标签。$ACC$ 可以定义为:

$$ACC = \frac{1}{n}\sum_{i=1}^{n}\delta(q_i, map(p_i))$$

其中:$n$ 是样本总数。如果 $x = y$,则 $\delta(x, y) = 1$;如果 $x \neq y$,则 $\delta(x, y) = 0$。而 $map(\cdot)$ 是置换映射函数,它将簇标签映射到类标签。最佳映射可以通过 Kuhn-Munkres 算法获取。$ACC$ 是 0 ~ 1 的值,$ACC$ 的值越大说明聚类效果越好。

2) 归一化互信息(Normalized Mutual Information, NMI)。$NMI$ 是一种外部评价标准,它用来评价算法在一个数据集上的聚类结果与该数据集真实划分的相似程度。用 $C$ 表示真实标签中类的集合,用 $C'$ 表示聚类算法获得的簇的集合。它们的互信息定义为:

$$MI(C, C') = \sum_{c_i \in C, c'_j \in C'} p(c_i, c'_j)\,\mathrm{lb}\,\frac{p(c_i, c'_j)}{p(c_i)\,p(c'_j)}$$

其中:$p(c_i)$ 和 $p(c'_j)$ 分别是样本属于类 $c_i$ 和簇 $c'_j$ 的概率;$p(c_i, c'_j)$ 是样本同时属于类 $c_i$ 和簇 $c'_j$ 的概率。实验中使用下面的归一化互信息:

$$NMI(C, C') = \frac{MI(C, C')}{\max(H(C), H(C'))}$$

其中:$H(C)$ 和 $H(C')$ 分别是类 $C$ 和簇 $C'$ 对应的信息熵。容易验证 $NMI$ 位于 0 ~ 1,并且 $NMI$ 的值越大说明聚类效果越好。

3) 聚类纯度($Purity$) 是一种简单的聚类评价方法,只需计算正确聚类的样本数占样本总数的比例,其计算方法如下:

$$purity = \frac{1}{n}\sum_{k}\max(c'_k, c_j)$$

其中:用 $C = \{c_1, c_2, \cdots, c_k\}$ 表示真实标签中类的集合;用 $C' = \{c'_1, c'_2, \cdots, c'_k\}$ 表示聚类算法获得的簇的集合。$Purity$ 同样位于 0 ~ 1,并且 $Purity$ 的值越大说明聚类效果越好。

### 3.4 结果与分析

表 2 ~ 4 分别列出了不同的聚类方法在这些数据集上聚类准确性、归一化互信息和聚类纯度的结果。

实验结果表明多核方法(Coreg、RMKKM 和 GMKCF)普遍优于单核方法(KKM、SC 和 KCF)。从表 2 聚类准确性指标可看出多核方法在多个数据集上的平均结果达到 0.5809,而单核方法的平均结果为 0.4915,多核方法在聚类准确性上的平均提升达到了 18.2%;从表 3 归一化互信息指标可看出多核方法在多个数据集上的平均结果达到 0.3741,而单核方法的平均结果为 0.2463,多核方法在归一化互信息上的平均提升达到了 51.8%;从表 4 聚类纯度指标可看出多核方法在多个数据集上的平均结果达到 0.6599,而单核方法的平均结果为 0.5766,多核方法在归一化互信息上的平均提升达到了 14.4%。

实验结果表明本文提出的多核概念分解方法要优于其他单核方法和多核方法。三种不同指标上 GMKCF 在多个数据集上的平均结果明显高于其他方法。

具体来看,GMKCF 在聚类准确性上达到 0.6145,而第二名的算法 Coreg 为 0.5664,性能提升为 8.5%。GMKCF 在归一化互信息上达到 0.4344,第二名为 0.4032,性能提升为 7.7%;GMKCF 在聚类纯度上达到 0.6982,第二名为 0.6756,性能提升为 3.3%。

表 2 各聚类算法的聚类准确性(ACC)对比

Tab. 2 Comparison of ACC among different algorithms

| 数据集 | KKM | SC | KCF | Coreg | RMKKM | GMKCF |
|---|---|---|---|---|---|---|
| BBC | 0.4567 | 0.4062 | 0.4701 | 0.5604 | 0.4707 | **0.6132** |
| TR31 | 0.4376 | 0.4789 | 0.423 | 0.4363 | **0.4969** | 0.4759 |
| K1B | 0.6422 | 0.554 | 0.6145 | 0.6962 | 0.6947 | **0.7694** |
| WebKB | 0.5045 | 0.4438 | 0.4669 | 0.5728 | 0.5859 | **0.5995** |
| 平均值 | 0.5103 | 0.4707 | 0.4936 | 0.5664 | 0.5620 | **0.6145** |

表 3 各聚类算法的归一化互信息(NMI)对比

Tab. 3 Comparison of NMI among different algorithms

| 数据集 | KKM | SC | KCF | Coreg | RMKKM | GMKCF |
|---|---|---|---|---|---|---|
| BBC | 0.1961 | 0.2018 | 0.2378 | 0.3469 | 0.2269 | **0.3977** |
| TR31 | 0.2029 | 0.3436 | 0.2506 | 0.3292 | 0.2628 | **0.3540** |
| K1B | 0.2815 | 0.3916 | 0.3512 | 0.564 | 0.3325 | **0.6204** |
| WebKB | 0.1981 | 0.1132 | 0.1872 | **0.3727** | 0.3169 | 0.3656 |
| 平均值 | 0.2197 | 0.2626 | 0.2567 | 0.4032 | 0.2848 | **0.4344** |

表 4 各聚类算法的聚类纯度(purity)对比

Tab. 4 Comparison of purity among different algorithms

| 数据集 | KKM | SC | KCF | Coreg | RMKKM | GMKCF |
|---|---|---|---|---|---|---|
| BBC | 0.4840 | 0.4675 | 0.519 | 0.5861 | 0.5046 | **0.6436** |
| TR31 | 0.5204 | 0.6284 | 0.5522 | 0.6083 | 0.5622 | **0.6309** |
| K1B | 0.7164 | 0.7702 | 0.7457 | 0.8298 | 0.7336 | **0.8539** |
| WebKB | 0.5349 | 0.4516 | 0.5288 | **0.6780** | 0.6228 | 0.6642 |
| 平均值 | 0.5639 | 0.5794 | 0.5864 | 0.6756 | 0.6058 | 0.6982 |

需要指出的是多核方法 Coreg 和 RMKKM 都带有超参数,无监督聚类问题中如何选择有效的超参数本身就是一个非常困难的问题。而本文提出的 GMKCF 算法无需设置其他特定参数,极大提升了算法的实际可用性。

此外,本文提出的 GMKCF 算法在空间复杂度上和其他

多核方法类似，都是 $O(n^2)$，从时间复杂度看 GMKCF 和 RMKKM 都是 $O(n^2)$，而 Coreg 的时间复杂度为 $O(n^3)$；并且 GMKCF 和 RMKKM 中主要涉及矩阵和向量的基本操作，可以借助 MapReduce 等框架容易实现分布式部署，而 Coreg 由于需要计算特征空间导致分布式实现较为困难。

实验结果表明，本文提出的多核概念分解方法在多种聚类评价指标上的结果要优于其他单核和多核聚类方法，无需设置超参数，并且算法复杂度较低，容易分布式部署。

## 4 结语

针对核概念分解模型在实际应用中面临的核函数选择问题，本文提出基于多核全局融合的概念分解模型。与核概念分解模型类似，本文推导出对应的迭代式乘法更新公式作为求解算法并且证明算法的收敛性。多个基准数据集上的实验结果表明，本文算法在不引入额外超参数的情况下能够有效提升核分解模型在实际应用中的聚类性能。未来，我们将进一步研究如何在分布式环境中部署实施多核概念分解算法。

参考文献(References)

This work is partially supported by the National Natural Science Foundation of China (61502289).

**LI Fei**, born in 1993, M. S. candidate. His research interests include data mining, machine learning.

**DU Liang**, born in 1985, Ph. D., lecturer. His research interests include data mining, machine learning.

**REN Chaohong**, born in 1994, M. S. candidate. His research interests include data mining, machine learning.